\documentclass{article}

\PassOptionsToPackage{numbers}{natbib}

\usepackage[final]{nips_2018}

\usepackage[utf8]{inputenc} 
\usepackage[T1]{fontenc}    
\usepackage{hyperref}       
\usepackage{url}            
\usepackage{booktabs}       
\usepackage{amsfonts}       
\usepackage{nicefrac}       
\usepackage{microtype}      
\usepackage{tabu}
\usepackage{amsmath}
\usepackage{tabularx}
\usepackage{colortbl}
\usepackage{hhline}
\usepackage{graphicx}
\usepackage{natbib}
\usepackage{subcaption}
\title{Online Collective Animal Movement Activity Recognition}

\author{
  Kehinde~Owoeye\\ 
  Department of Computer Science\\
  University College London\\
  London, WC1E 6EA \\
  \texttt{ucabowo@ucl.ac.uk} \\
  \And 
  Stephen~Hailes\\
  Department of Computer Science\\
  University College London\\
  London, WC1E 6EA \\
  \texttt{s.hailes@ucl.ac.uk} \\
}
\begin{document}

\maketitle

\begin{abstract}

Learning the  activities of animals  is important for the purpose of monitoring their welfare vis a vis their behaviour with respect to their environment and conspecifics. While previous works have largely focused on activity recognition in a single animal, little or no work has been done in learning the collective behaviour of animals. In this work, we address the problem of recognising the collective movement activities of a group of sheep in a flock. We present a discriminative framework that learns to track the positions and velocities of all the animals in the flock in an online manner whilst estimating their collective activity. We investigate the performance of two simple deep network architectures and show that we can learn the collective activities with good accuracy even when the distribution of the activities is skewed.
\end{abstract}


\section{Introduction}

Recognising the collective movement  activities of animals  is important for several real world applications such as monitoring their welfare vis a vis their behaviour with respect to their environment and conspecifics, predicting the onset of an epidemic or attacks from predators most especially for animals who live in cooperative societies. While several works have been carried out in recognising collective activities with respective to human interactions~\citep{choi2012unified, wang2017recurrent}, previous works in the animal behaviour community however have either  tried to recognize collective behaviour from different species of animal  where a model is fed the entire collective behaviour input and asked to identify the specie generating such behaviour~\citep{delellis2014collective} or learn the activity of just a single animal~\citep{kamminga2017generic,grunewalder2012movement}. There are however  problems with these approaches. Learning collective activity in an offline manner doesn't scale into real world applications where the collective activity is required real time for example in monitoring poaching activities. In addition, learning the individual activity and aggregating them is computationally intensive and may not be informative compared to the collective context. 

 With the recent developments in deep learning research however, there has been a rise in developing models that can overcome some of these limitations. Notable among these techniques are the Convolution Neural Networks (CNNs) and Recurrent Neural Networks (RNNs) for modelling spatial and temporal dependencies respectively. In this work, we focus on the task of recognising the collective movement activities of a group of sheep in a flock. We gather movement data of 36 sheep and  investigate  the use of several deep learning models and two features (spatial orientations and velocities) of the sheep to capture their collective activities. Our approach is one that learns in an online fashion the collective activity of the flock using a deep recurrent and convolutional neural network.

The main contributions of this paper are: (i) We extend the idea of activity recognition in animal behaviour  to the collective settings and  propose a model that can learn the collective movement behaviour in an online fashion (ii) We demonstrate that our approach can learn to classify activities even when some activities are underrepresented.

\begin{figure}[h]
\begin{minipage}[b]{.5\textwidth}
\centering\includegraphics[width=7cm]{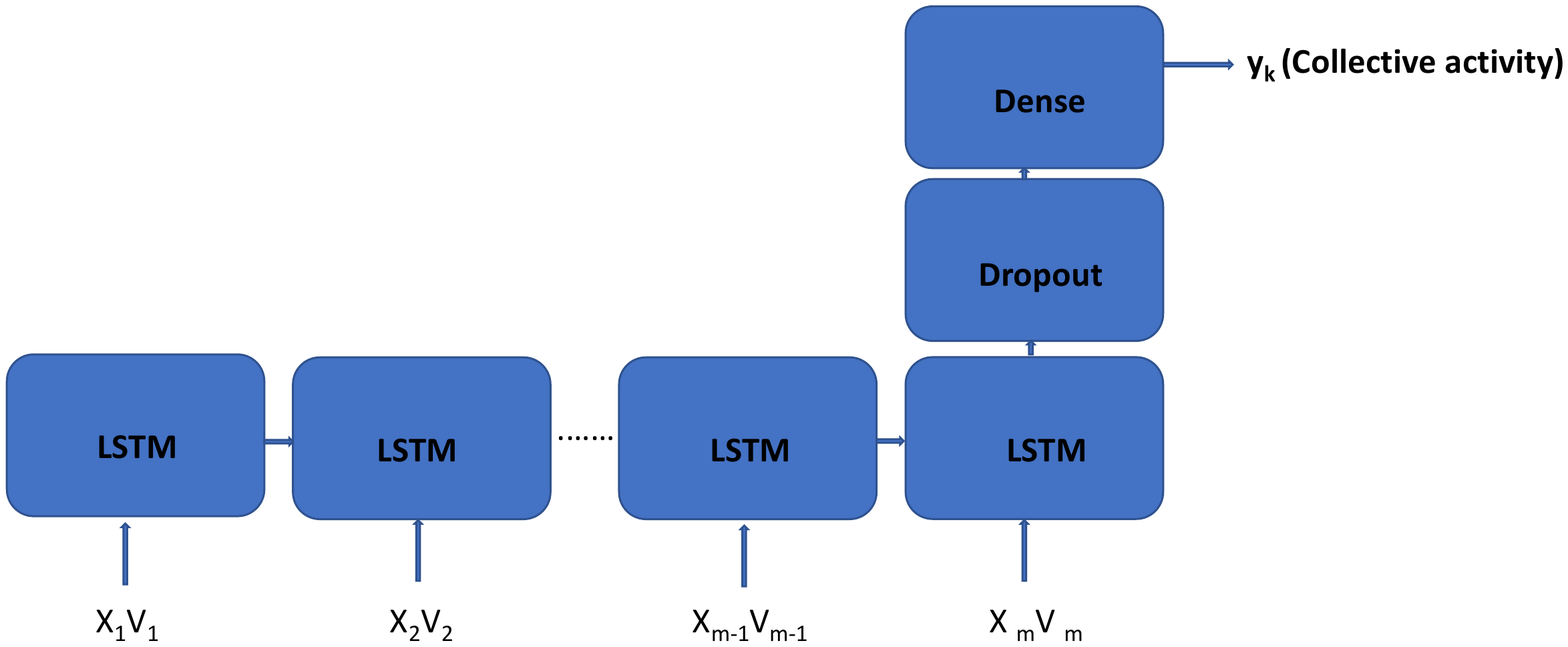}
\end{minipage}%
\begin{minipage}[b]{.5\textwidth}
\centering\includegraphics[width=8cm]{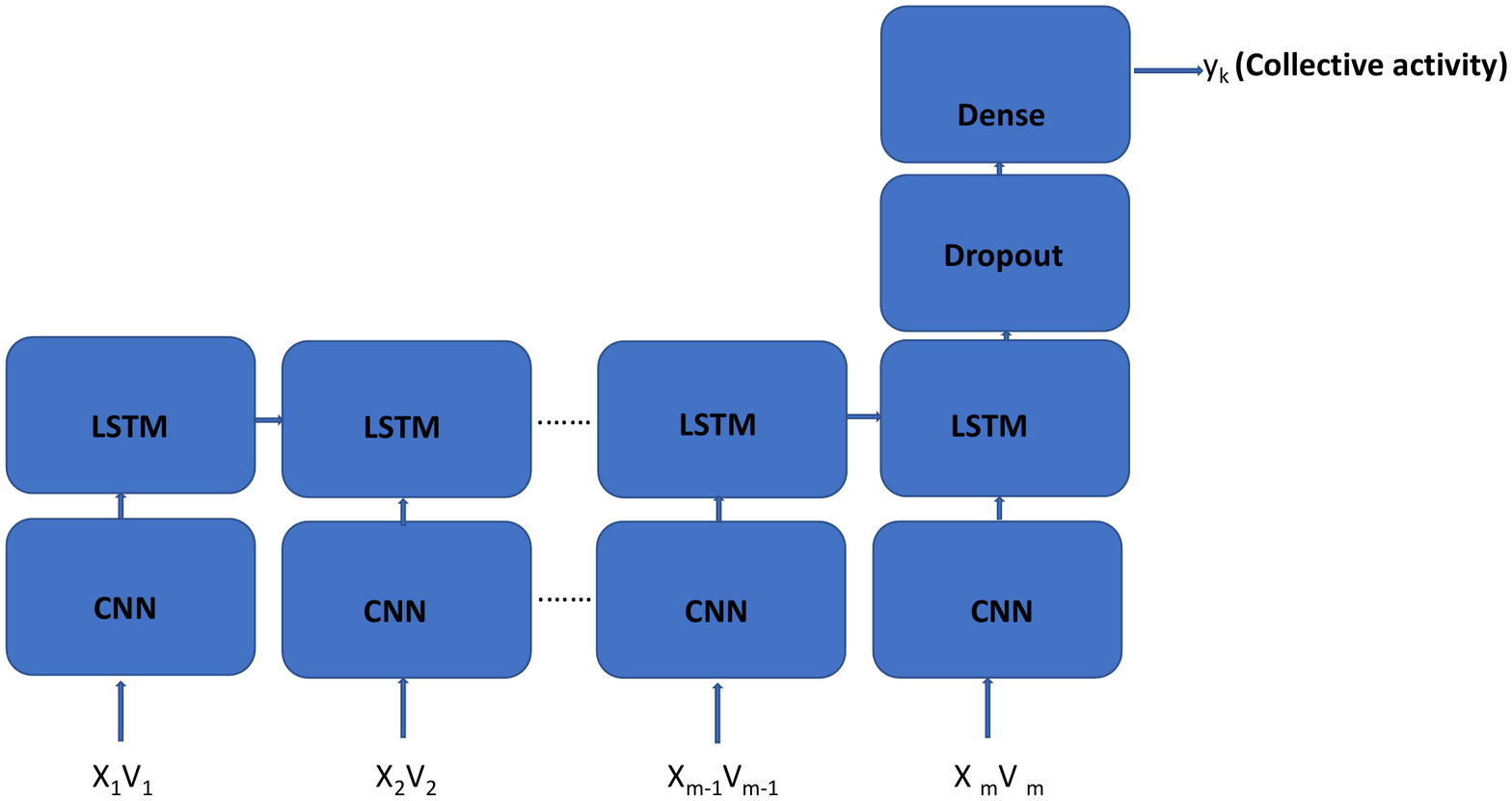}
\end{minipage}%
\caption{The basic LSTM and CNN+LSTM architectures on the left and right respectively. The inputs are $X_{m}V_{m}$ (orientation in space and velocity respectively) where $m$ represents the size of the temporal window. Only the last output of both architectures is selected as the predicted activity (many to one architecture).}
\label{fig:1}
\end{figure}

\section{Problem Formulation}
We briefly describe the formulation of the collective movement activity recognition problem here.  Assume $D_{tr} = \{ (X_{n}, V_{n}), y_{k} \}$ represents the training dataset of $N_{s}$ samples and $D_{te} = \{ (X_{n}, V_{n}) \}$ denotes the test set of $N_{d}$ samples, where  $N_{d}$  may or may not be equal to $N_{s}$, and $X_{n}, V_{n}$ represents spatial and velocity features of all the animals of interest respectively, while the corresponding label space representing the collective activities are $Y = \{y_{1},y_{2},....,y_{k} \}$ with $k$ being the number of unique collective activities. Now the problem is as follows, given a history of observations over a time window $t$ described by $\{X_{i}, V_{i}|i =
1, ..., t\}$, the goal is to predict $y_{k}  \in Y $ for each datapoint in $D_{te}$. That is, we aim to learn  $p(y_{tk} | (X_{t}, V_{t}),(X_{t-1}, V_{t-1}),.....,(X_{t-m}, V_{t-m}))$
 where $m$ is the time window of feature observations used for the prediction task.

\section{Model Description}

\textbf{Temporal Dynamics Modelling:} We use the recurrent neural network (RNN) to model temporal dependencies. In particular, we use the LSTM (Long Short-Term Memory)~\citep{hochreiter1997long} a variant of the RNN to model long term dependencies and has been used in previous studies~\citep{graves2013generating, sutskever2014sequence} for handwriting synthesis and language translation. The LSTM can be described by the following equations:

\begin{equation}
\begin{aligned}
i_{t} &=  \sigma (W_{xi}x_{t} + W_{hi}h_{t-1} + W_{ci}C_{t-1} + b_{i})\\
f_{t} &=\sigma(W_{xf}x_{t} +W_{hf}h_{t-1} +W_{cf}C_{t-1} +b_{f})\\
C_{t} & = f_{t}C_{t-1} + i_{t} tanh (W_{xc}x_{t} + W_{hc} h_{t-1} + b_{c})\\
o_{t} & = \sigma(W_{xo}x_{t} + W_{ho}h_{t-1} + W_{co}c_{t} + b_{o})\\
h_{t} & = o_{t} tanh(C_{t})
\end{aligned}
\end{equation}

where $i_{t}$ , $f_{t}$ , and $o_{t}$ are input, forget, and output gate respectively. $W$ is the weight matrix, $x_{t}$ is the current input data, $h_{t-1}$ is previous hidden output, $C_{t}$ is the cell state and $\sigma$ is the logistic sigmoid function. 

\textbf{Temporal and Spatial Dynamics Modelling:} We use the recurrent convolutional network model of~\citep{donahue2015long} where the input is passed through a CNN to produce a reduced representation of the input which is further passed into a RNN (Figure~\ref{fig:1}).


\section{Data Acquisition and Pre-processing}
\textbf{Data collection:} We present here our data collection system. All experiments involved with the animals complied with the Australian ethics laws guiding the handling of animals. We collected movement data of 36 sheep in a field with the aid of a GPS device designed in house for two days and attached to the back of the sheep. Previous work has shown that such harness equipment carried by sheep does not affect their locomotion~\citep{hobbs2012data}. Data was collected at a sampling rate of 1 sample/s to ensure all forms of interesting movement patterns by the sheep are captured. Each dataset contains a phase in which loggers were attached to each sheep in a holding pen followed by a phase in which the sheep were herded into the field and then a phase in which the sheep were left to roam across the field and a final phase in which the sheep were herded back to the holding pen to have the logging device removed and re-charged. We extracted only portions of the dataset where all the loggers were working. All missing data were interpolated between the last and next seen co-ordinates using the expectation-maximization algorithm~\citep{dempster1977maximum}. 

\textbf{Data labelling:} With the aid of a viewer designed in house, we label the collective activities with respect to the instances where they occur in the dataset. Movement data of all the sheep in the flock were labelled with respective to the collective behaviours described in (Table~\ref{tab:1}). Due diligence was ensured to make sure the labelling was of high quality.

\begin{table}[h!]
\caption{Observed collective activities and their description.}
 \begin{tabu} to 1\textwidth {  X[l]  X[l] }
\toprule
Collective Movement Activities & Description  \\
\bottomrule
Not Active & Where the animals are gathered together in close proximity with little or no movement activities. Correspond to instances where the animals are resting or sleeping. \\
\midrule
Active & The animals are moving and  scattered in their habitat with diverse movement activities.  \\
\midrule
Herd Movement & The animals are being herded at a high velocity over a narrow space.\\
 \bottomrule
\end{tabu}
\label{tab:1}
\end{table}

\addtolength{\tabcolsep}{14pt} 
\begin{table}[h!]
\begin{center}
\caption{Number of samples for  each collective activity in the training and test datasets. The distribution can be seen to be highly skewed with the Herd movement mostly affected.}
\begin{tabular}{c c c c}
\toprule
Dataset & Not Active & Active & Herd Movement\\
 \midrule
Train & 21801 (37.55\%) &  35811 (61.68\%) & 452 (0.78\%) \\
 \midrule
Test & 26718 (41.96\%) & 36355 (56.24\%) & 597 (0.94\%) \\
 \bottomrule

\end{tabular}

\label{tab:2}
\end{center}
\end{table}

\section{Experiments \& Results}

\textbf{Experiments:} We used dataset of activities for one day for training and another day for testing see (Table~\ref{tab:2}). Models were trained with the following parameters: Adam optimizer~\citep{kingma2014adam} at a learning rate = 0.001, look-back \& recurrent cells = 30, batch-size=10 over 50 epochs, dropout = 0.2, categorical crossentropy for all losses and a softmax for all classification tasks, activation = tanh. For the CNN segment of the CNN+LSTM model, we used a 1D CNN with a  2x2 filter, stride =1  and relu activation. All models were  basic with only one layer and trained on a 2.3 GHz Intel Core i5 PC.
We investigate the use of two features, the velocities of the animals and the distance of each sheep to their centroid in an ablation manner with respect to the two models. The collective activities were one hot encoded. All evaluations were carried out with respect to the classification accuracy and confusion matrix. 

\textbf{Results:} From the results in (Table~\ref{tab:3}), the LSTM (velocities \& distance to centroid) architecture outperform others although marginally relative to the LSTM (velocities) and CNN+LSTM (velocities \& distance to centroid) architectures. The distance of each sheep to the entire sheep centroid seems to be the least informative of the two features but produces better results when combined with the velocities of the sheep using LSTM.  On the average the CNN+LSTM architectures perform slightly better than the LSTM architectures over all feature combinations. More impressive is the fact that our model is able to classify very well the underrepresented activity. As seen in the confusion matrices (Figure~\ref{fig:2}), only the models including spatial features were able to learn the Herd movement activity with the CNN+LSTM performing better than the LSTM while the remaining two models\footnote{We omit the remaining two models here for brevity.} entirely misclassified this activity. This suggests that a fusion of spatial and movement features is essential to disentangle some of these complex collective activities especially in very rough and challenging terrains.

\begin{table}[h]
\begin{center}
\caption{Classification accuracy averaged over fifty epochs. Results show that the LSTM model  with both spatial and velocity features gives better classification accuracy .}
\begin{tabular}{c c}
\toprule
Models \& Features  & Classification Accuracy (\%) \\
 \midrule
LSTM (velocities)& $75.62 \pm 0.25$ \\
LSTM (distance to centroid)& $ 60.80 \pm 1.91 $ \\
\textbf{LSTM (velocities \& distance to centroid)} &  $\textbf{77.02} \pm \textbf{1.11} $\\
CNN+LSTM (velocities)& $73.65 \pm 0.33$ \\
CNN+LSTM (distance to centroid)& $70.63 \pm 1.37$ \\
{CNN+LSTM (velocities \& distance to centroid)} & ${74.51} \pm {1.03}$ \\
\bottomrule
\end{tabular}
\label{tab:3}
\end{center}
\end{table}


\section{Conclusions \& Future Work}
In this paper, we have shown how to learn the collective movement activities of sheep using deep neural network. Our approach leveraged the fusion of spatio-temporal features from all animals of interest to learn collective activities even when their distribution is skewed. This work has implications for example in building automatic systems that can help farmers better understand the health of the flock as a whole for further applications in epidemic management as well as help conservationist learn collective behaviours of animals that are indicative of poaching activities. It is however not clear how our approach will perform when faced with a flock of different size. One potential solution is to use an embedding to project behaviours (features) into a low dimensional space similar to~\citep{tang2017latent}. While we have used animals who live in co-operative societies, the method used here may not scale into other animal societies without any structure.  In the future, we aim to explore the use of an hierarchical model to improve the classification accuracy most especially when the distribution of activities is skewed. While we have used very simple neural network architectures, in the future, it is our desire  to investigate other complex deep learning architectures to improve the accuracy.
\medskip

\addtolength{\tabcolsep}{-18pt} 
\begin{figure}[h]

\begin{minipage}[b]{.5\textwidth}
\centering
\begin{tabularx}{2\textwidth}{>{\bfseries}c|c c c |}
 & \textbf{Herd M.} & \textbf{Active} & \multicolumn{1}{c}{\textbf{Not Active}} \\
\hhline{----}
Herd M. & 0.84 \cellcolor[gray]{.8}& 0.16 & 0  \\
Active & 0 & 0.46 \cellcolor[gray]{.8}& 0.54 \\
Not Active & 0 & 0.03 & 0.97 \cellcolor[gray]{.8} \\
\hhline{~---}
\end{tabularx}
\subcaption{CNN+LSTM (velocities \& dist. to centroid)}\label{}
\end{minipage}
\begin{minipage}[b]{.5\textwidth}
\centering
\begin{tabularx}{.7\textwidth}{>{\bfseries}c|c c c |}
 & \textbf{Herd M.} & \textbf{Active} & \multicolumn{1}{c}{\textbf{Not Active}} \\
\hhline{----}
Herd M. & 0 \cellcolor[gray]{.8}& 1 & 0  \\
Active & 0 & 0.57 \cellcolor[gray]{.8}& 0.43  \\
Not Active & 0 & 0.13 & 0.87 \cellcolor[gray]{.8} \\
\hhline{~---}
\end{tabularx}
\subcaption{CNN+LSTM (velocities)}\label{}
\end{minipage}\\[2ex]

\begin{minipage}[b]{.5\textwidth}
\centering
\begin{tabularx}{2\textwidth}{>{\bfseries}c|c c c |}
 & \textbf{Herd M.} & \textbf{Active} & \multicolumn{1}{c}{\textbf{Not Active}} \\
\hhline{----}
Herd M. & 0.73 \cellcolor[gray]{.8}& 0.27 & 0  \\
Active & 0 & 0.48 \cellcolor[gray]{.8}& 0.52  \\
Not Active & 0 & 0.03 & 0.97 \cellcolor[gray]{.8} \\
\hhline{~---}
\end{tabularx}
\subcaption{LSTM (velocities \& dist. to centroid)}
\end{minipage}
\begin{minipage}[b]{.5\textwidth}
\centering
\begin{tabularx}{0.7\textwidth}{>{\bfseries}c|c c c |}
 & \textbf{Herd M.} & \textbf{Active} & \multicolumn{1}{c}{\textbf{Not Active}} \\
\hhline{----}
Herd M. & 0 \cellcolor[gray]{.8}& 1 & 0  \\
Active & 0 & 0.64 \cellcolor[gray]{.8}& 0.36  \\
Not Active & 0 & 0.15 & 0.85 \cellcolor[gray]{.8} \\
\hhline{~---}
\end{tabularx}
\subcaption{ LSTM (velocities)}
\end{minipage}\\[2ex]

\caption{Confusion matrix  for the top four performing models. The models where the spatial features were included show a significantly higher accuracy with respect to the underrepresented Herd movement activity.}
\label{fig:2}
\end{figure}

\clearpage

\bibliography{references}

\end{document}